\PassOptionsToPackage{table,dvipsnames}{xcolor}
\documentclass[]{selfevolagent}

\usepackage{microtype}
\microtypesetup{expansion=false}
\usepackage{amsmath}
\usepackage{amsfonts}
\usepackage{xcolor}
\usepackage{graphicx}
\usepackage{booktabs}
\usepackage{colortbl}
\usepackage{array}
\usepackage{placeins}
\usepackage{wrapfig}
\definecolor{overallblue}{RGB}{241,247,252}

\newcommand{\ours}{\texttt{OPOD}}
\newcommand{\tworowhead}[1]{\raisebox{-0.70\normalbaselineskip}[0pt][0pt]{#1}}
\newcommand{\benchhead}[1]{{\fontsize{7.6}{8.2}\selectfont #1}}
\DeclareRobustCommand{\mailicon}{\raisebox{-0.12em}{\includegraphics[height=0.92em]{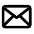}}}
\DeclareRobustCommand{\githubicon}{\raisebox{-0.14em}{\includegraphics[height=0.95em]{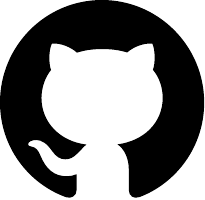}}}
\DeclareRobustCommand{\huggingfaceicon}{\raisebox{-0.17em}{\includegraphics[height=1.05em]{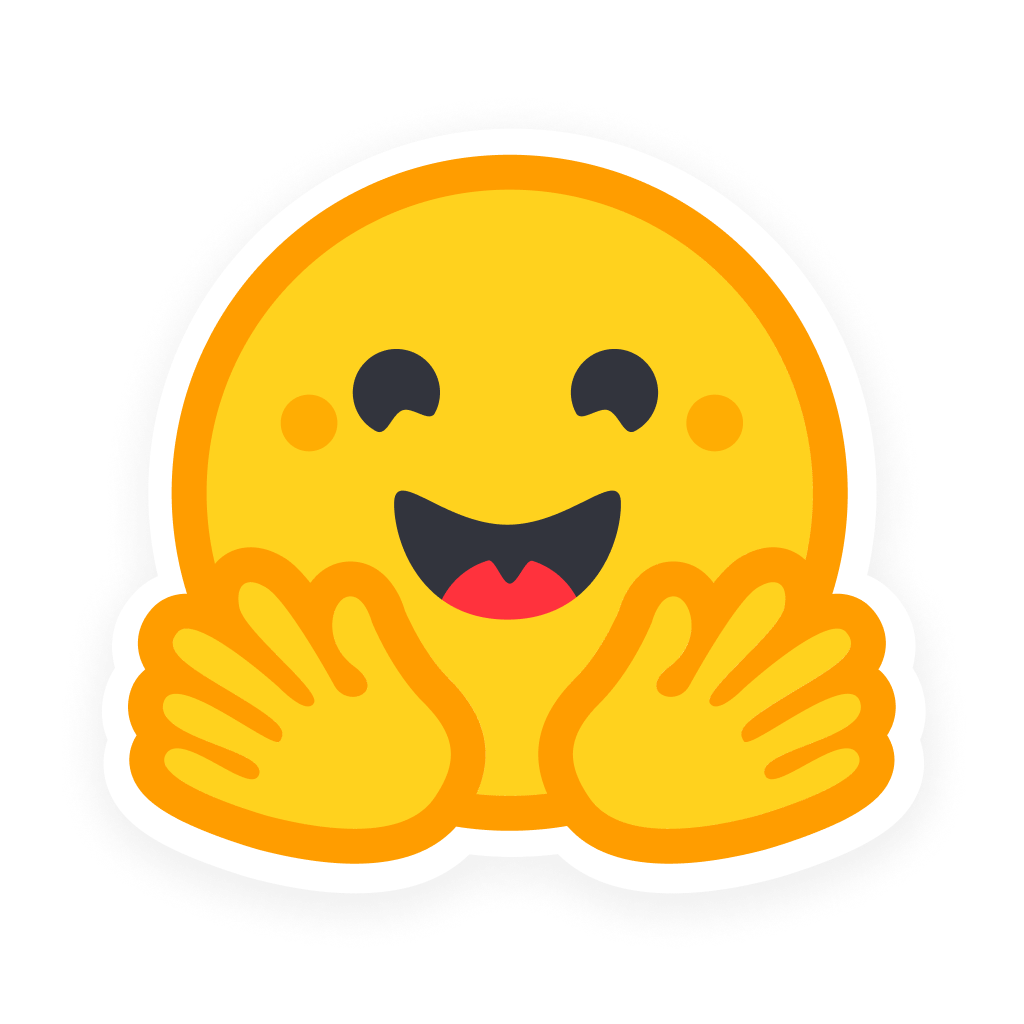}}}
\DeclareRobustCommand{\metadataicon}[1]{\makebox[1.20em][c]{#1}\hspace{0.08em}}
\providecommand{\Description}[1]{}

\title{OPOD: On-Policy Omni Distillation}

\author[1,*]{Tong Zhao}
\author[1]{Yuyang Hu}
\author[1]{Yutao Zhu}
\author[2]{Reed Li}
\author[2]{Haijin Liang}
\author[2]{Haibo Shi}
\author[2,\dagger]{Yu Lu}
\author[1,\dagger]{Zhicheng Dou}

\affiliation{\textbf{Affiliations:} $^1$Renmin University of China;\enspace $^2$Tencent}
\contribution{$^*$Work done during an internship at Tencent.\enspace
$^\dagger$Corresponding authors.}

\abstract{
Omni-modal models provide a unified interface for text, images, and audio. However, improving these abilities together remains difficult, as post-training on pooled multimodal data often fails to preserve the strengths of modality teachers. On-policy distillation (OPD) has recently become popular in model post-training. It samples responses from the current student and compares the teacher's and student's next-token distributions along those responses, yielding dense supervision while reducing the mismatch between training and inference. Despite these advantages, standard OPD does not readily extend to several modality teachers. Their guidance may favor conflicting changes to the shared model, while matching each teacher's next-token distribution can prevent the student from moving beyond that teacher. To address these challenges, we propose On-Policy Omni Distillation (\ours{}), which consolidates text, image, and audio teachers into one omni model. \ours{} routes each response to the corresponding teacher, controls the teachers independently, and applies guidance only when the teacher assigns a higher probability to the generated token. The selected teacher also evaluates answer confidence and whether the reasoning increases support for the answer. Extensive experiments on twelve benchmarks show that \ours{} achieves the best average at three model scales, reaching 70.8, 51.7, and 46.2 and outperforming the strongest comparator by 2.1, 1.8, and 1.7 points. At 30B, it surpasses the base model and pooled RL training on all twelve benchmarks, and ranks first or second on eleven even when the teachers are included. Only the student is retained for deployment.
}

\metadata[\metadataicon{\mailicon}Contact]{\email{zhaotong7@ruc.edu.cn}, \email{luyusearch@foxmail.com}, \email{dou@ruc.edu.cn}}
\metadata[\metadataicon{\githubicon}Code]{\href{https://github.com/VincentZhao2002/OPOD}{github.com/VincentZhao2002/OPOD}}
\metadata[\metadataicon{\huggingfaceicon}Models]{\href{https://huggingface.co/Tung111/OPOD-Qwen3-Omni-30B-A3B}{30B-A3B}; \href{https://huggingface.co/Tung111/OPOD-Qwen2.5-Omni-7B}{7B}; \href{https://huggingface.co/Tung111/OPOD-Qwen2.5-Omni-3B}{3B}}

\hypersetup{
  pdftitle={OPOD: On-Policy Omni Distillation},
  pdfauthor={Tong Zhao, Yuyang Hu, Yutao Zhu, Reed Li, Haijin Liang, Haibo Shi, Yu Lu, and Zhicheng Dou}
}

\begin{document}

\maketitle

\section{Introduction}

The rapid progress of multimodal pretraining and instruction tuning has enabled foundation models to process language, images, and audio through a unified interface~\citep{li2023blip2,dai2023instructblip,chu2024qwen2audio,qwen25omni2025,xu2025qwen3omni}. Despite these advances, a shared backbone does not guarantee that improvements learned from one modality will preserve performance on the others. Text, vision, and audio also require different forms of reasoning~\citep{wang2024mmlupro,yue2024mmmu,lu2024mathvista,sakshi2025mmau}, and reinforcement learning on their combined data can create performance tradeoffs across modalities~\citep{zhang2025mmrlhf}. This raises our central question: \emph{How can we combine expertise from different modalities in one model without compromising performance in any modality?}

\begin{figure*}[t]
\centering
\includegraphics[width=\textwidth]{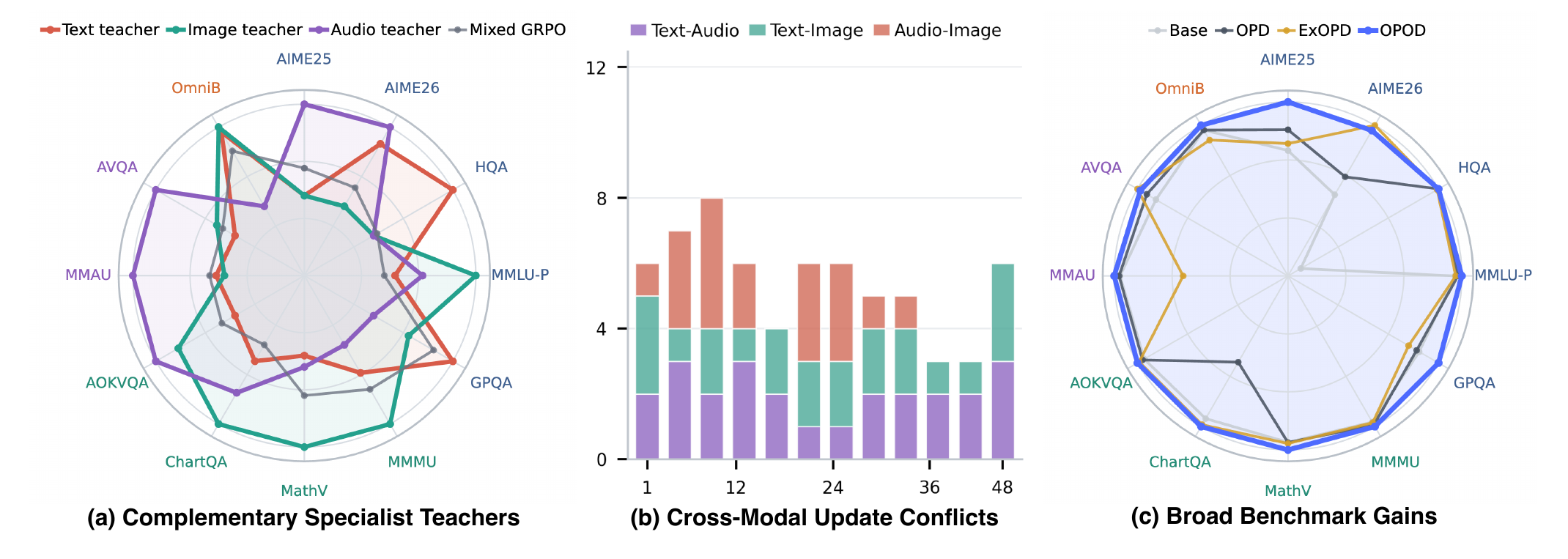}
\Description{Three panels compare modality specialists with pooled training, show conflicts among specialist update directions, and summarize benchmark gains from OPOD.}
\caption{Motivating observations for \ours{}. (a) Under the same GRPO setup, modality specialists develop complementary strengths that training on pooled data does not consistently recover. These specialists then serve as our teachers. (b) Their directions of parameter change from the common base frequently conflict. (c) \ours{} yields broader benchmark gains than the base model, native OPD, and ExOPD.}
\label{fig:intro}
\end{figure*}

A straightforward approach is to train one model with reinforcement learning on a mixture of text, image, and audio data. We examine this approach in a controlled experiment where all models are trained with GRPO~\citep{shao2024deepseekmath}. Starting from the same base model, we train one specialist for each modality and a pooled model on the combined data. Figure~\ref{fig:intro}(a) shows that the specialists develop complementary strengths, whereas the pooled model does not consistently match them. These specialists then serve as our teachers. Our next goal is to consolidate their complementary capabilities into one student without retaining an ensemble at inference. Knowledge distillation is a standard mechanism for this purpose because it transfers the predictive behavior of one or more teachers into a student~\citep{hinton2015distilling,you2017multipleteachers,wan2024knowledge}. Conventional sequence distillation, however, usually trains on fixed responses or responses generated by the teacher. In autoregressive generation, such trajectories may differ from those the student visits at inference. Recent model post-training methods therefore increasingly adopt on-policy distillation (OPD) to reduce this mismatch~\citep{gu2024minillm,agarwal2024onpolicy}. OPD samples a response from the current student and compares the teacher's and student's next-token distributions at each visited prefix. This provides dense supervision throughout the response, even when the final outcome reward is uninformative. Because the prefixes are sampled from the current student, the feedback remains aligned with its evolving policy. These properties provide a practical way to consolidate several specialists while retaining only the student for deployment.

However, extending OPD to several modality teachers introduces two challenges. First, different teachers may push the shared backbone in different directions. As shown in Figure~\ref{fig:intro}(b), their directions of parameter change from the common base frequently conflict. This makes a shared guidance weight ill suited to regulating the three signals independently. Second, a teacher may constrain the student even when the student is already better at particular tokens. Standard OPD minimizes their discrepancy at every visited prefix, even when the student assigns a higher likelihood to the sampled token. This symmetric objective can pull the student backward and impose a teacher ceiling~\citep{yang2026learning}. The first challenge concerns coordination among teachers, whereas the second concerns when a teacher should guide the student. Recent methods route or arbitrate among several OPD teachers~\citep{ma2026mopd,yin2026hopd}, but they do not address both challenges when consolidating text, image, and audio specialists into one shared model.

To address these challenges, we propose On-Policy Omni Distillation (\ours{}). Adaptive modality control assigns each modality its own constraint budget and guidance weight. This allows the influence of each teacher to change independently according to the current constraint signal. One-sided guidance applies a token constraint only when the selected teacher assigns a higher likelihood than the student. It preserves useful corrections without preventing the student from moving beyond its teachers. The selected teacher also verifies the final answer and assesses whether the reasoning supports it~\citep{lightman2024verify,setlur2025rewarding}. This provides additional feedback on the reasoning process without human annotations of intermediate steps or tree search. As shown in Figure~\ref{fig:intro}(c), native OPD and ExOPD improve selected capabilities, whereas \ours{} produces more consistent gains across modalities.

Our main contributions are threefold:
\begin{itemize}
    \item We identify conflicting guidance among modality teachers and the teacher ceiling as two key challenges when OPD consolidates several specialists into one model.
    \item We address these challenges with adaptive modality control and one-sided guidance, together with teacher verification of answers and reasoning.
    \item Extensive experiments on twelve benchmarks and three model scales show that \ours{} achieves the highest average, outperforming the strongest comparator by 2.1, 1.8, and 1.7 points at 30B, 7B, and 3B.
\end{itemize}

\section{Related Work}

\paragraph{On-policy distillation.}
Knowledge distillation transfers teacher behavior through softened predictions or teacher-generated sequences~\citep{hinton2015distilling,kim2016sequence,sanh2019distilbert}. For autoregressive models, these fixed trajectories can differ from the student's inference distribution. OPD instead samples from the student and computes teacher feedback on the resulting trajectories~\citep{agarwal2024onpolicy,gu2024minillm}. Recent work has expanded OPD in several directions, including reward extrapolation for learning beyond the teacher~\citep{yang2026learning}, black-box OPD without teacher logits~\citep{ye2025blackbox,wang2026prism}, analyses of OPD training dynamics~\citep{li2026rethinking}, and data balancing on both the student side and the teacher side~\citep{hou2026uniopd}. These methods strengthen OPD as a post-training recipe, while \ours{} extends it to coordinated modality specialists that constrain one shared omni-modal backbone during joint post-training.

\paragraph{Omni-modal post-training.}
Traditional multimodal training builds generalist models through paired pretraining and supervised instruction tuning, where an LLM is aligned with visual or audio encoders using large-scale curated data~\citep{li2023blip2,liu2023llava,dai2023instructblip,bai2023qwenvl,chu2024qwen2audio}. Recent omni-modal models further extend this recipe to unified text, image, audio, and video interfaces~\citep{qwen25omni2025,li2026omnigaia}. The field is now shifting from supervised alignment to post-training methods that improve reasoning and task reliability, including multimodal RL pre-alignment, on-policy distillation for video grounding, and offline preference optimization for omni-modal models~\citep{wang2026prism,li2026videoopd,li2026omnigaia,zhang2025mmrlhf,yuan2026visionopd}. These works show the importance of post-training for multimodal agents, while \ours{} targets joint text, image, and audio consolidation through modality-decoupled on-policy learning in a unified student.

\section{On-Policy Omni Distillation}
\subsection{Preliminaries}

We consider a student policy $\pi_\theta$ with a shared omni-modal backbone. The training data consist of prompts from different modalities, denoted as $(x_i, m_i)$, where $m_i \in \{\mathrm{text}, \mathrm{audio}, \mathrm{image}\}$. For each modality $m$, we assume a specialized teacher policy $\pi_T^m$. At each training iteration, the student samples a response $y_i = (y_{i,1}, \ldots, y_{i,T_i})$ of length $T_i$ from $\pi_\theta(\cdot \mid x_i)$, and the teacher selected by $m_i$ evaluates the same student-generated trajectory.

On-policy distillation optimizes the student on its own sampled responses. Given a prefix $(x_i, y_{i,<t})$, the teacher provides token-level supervision by comparing the log probability assigned by the teacher and the student:
\begin{equation}
\label{eq:opd-reward}
r_{i,t}^{\mathrm{opd}}
=
\log \pi_T^{m_i}(y_{i,t} \mid x_i, y_{i,<t})
-
\log \pi_\theta(y_{i,t} \mid x_i, y_{i,<t}).
\end{equation}
This log-ratio corresponds to the dense token-level reward induced by reverse KL on student rollouts.

\subsection{Overview}

\begin{figure}[!t]
\centering
\includegraphics[width=\columnwidth]{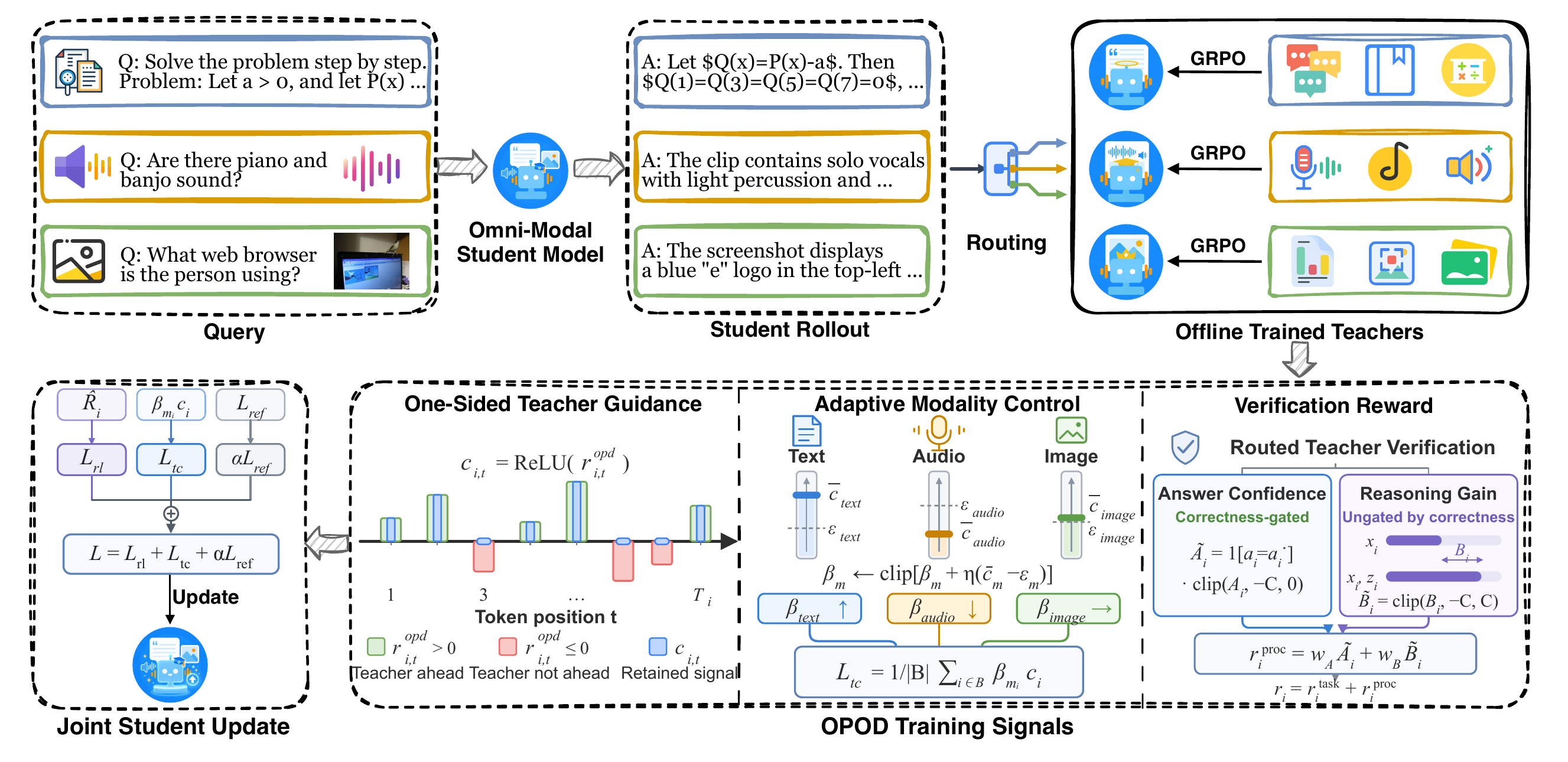}
\Description{An overview of OPOD showing modality routing, one-sided token guidance, modality-specific constraint control, teacher verification, and the shared student update.}
\caption{Overview of \ours{}. Student rollouts are routed by input modality to the corresponding teacher trained offline with GRPO. \ours{} combines one-sided token guidance, modality-specific constraint control, and teacher-based verification through answer confidence and reasoning gain. The resulting policy loss, teacher constraint, and reference regularizer jointly update the shared student.}
\label{fig:method}
\end{figure}

Figure~\ref{fig:method} summarizes the training pipeline of \ours{}. Given a query $x_i$ from modality $m_i$, the omni-modal student samples a rollout $y_i$. The known modality label $m_i$ routes this rollout to the corresponding offline teacher $\pi_T^{m_i}$. Each teacher is obtained by applying GRPO to its modality-specific data and evaluates the student-generated trajectory rather than producing a target response.

\ours{} converts this evaluation into three complementary components. One-Sided Teacher Guidance applies only where the routed teacher is ahead and stops once the student matches or exceeds it. Adaptive Modality Control maintains a separate constraint budget and dual weight for each modality, producing the modality-controlled teacher loss $\mathcal{L}_{\mathrm{tc}}$. Verification Reward reuses the routed teacher as a trajectory verifier and combines correctness-gated answer confidence with ungated reasoning gain. The verification reward is added to the task reward to form the rollout advantage used by $\mathcal{L}_{\mathrm{rl}}$. Finally, $\mathcal{L}_{\mathrm{rl}}$, $\mathcal{L}_{\mathrm{tc}}$, and the reference regularizer $\alpha\mathcal{L}_{\mathrm{ref}}$ are jointly optimized to update the shared student, while the offline teachers remain fixed. The teachers are discarded after post-training, so inference retains a single student without teacher-ensemble latency or memory.

\subsection{One-Sided Teacher Guidance}

For each sample $(x_i,m_i)$, the modality label selects the routed teacher $\pi_T^{m_i}$. We use the OPD reward $r_{i,t}^{\mathrm{opd}}$ defined above as a teacher-student margin on the sampled token. A positive margin means that the routed teacher assigns a higher probability than the student, while a negative margin means that the student already assigns a higher probability. Instead of treating both sides symmetrically, \ours{} keeps only the positive part:
\begin{equation}
\label{eq:one-sided}
c_{i,t} = \mathrm{ReLU}\left(r_{i,t}^{\mathrm{opd}}\right).
\end{equation}
Thus, teacher guidance is applied only where the teacher is ahead of the student. Once the student matches or exceeds the teacher on a token, the constraint becomes zero and no longer pulls the student back. We summarize the remaining teacher guidance on a rollout by
\[
c_i = \frac{1}{T_i}\sum_{t=1}^{T_i} c_{i,t}.
\]
This one-sided form preserves guidance on weak tokens while leaving room for the student to improve beyond saturated teachers.

\subsection{Adaptive Modality Control}

After obtaining the one-sided constraint, \ours{} controls its strength at the modality level. This is important because text, audio, and image teachers can progress at different speeds during on-policy training. We therefore maintain a separate trust-region weight $\beta_m$ for each modality.

For a mini-batch $\mathcal{B}$, let $\mathcal{B}_m = \{i \in \mathcal{B}: m_i=m\}$ be the subset of samples from modality $m$. For each modality represented in the mini-batch, we compute its average routed constraint as
\[
\bar{c}_m
=
\frac{1}{|\mathcal{B}_m|}
\sum_{i \in \mathcal{B}_m} c_i.
\]
Each modality has a target constraint budget $\epsilon_m>0$. We estimate this budget once from the mean warm-up constraint for that modality, lower-bounded by a shared floor $\epsilon_{\min}$, and then keep it fixed. The trust-region weight is updated as
\begin{equation}
\label{eq:beta-update}
\beta_m
\leftarrow
\mathrm{clip}\left(
\beta_m + \eta(\bar{c}_m - \epsilon_m),
\beta_{\min},
\beta_{\max}
\right),
\end{equation}
where $\eta>0$ is the dual step size and $0\leq\beta_{\min}<\beta_{\max}$ are the lower and upper bounds on the dual weight. This update is applied only when $|\mathcal{B}_m|>0$. When the student is far from a modality teacher, $\bar{c}_m$ exceeds the target budget and the corresponding weight increases; when the constraint is already small, the weight decreases. The current weights define the routed teacher constraint
\[
\mathcal{L}_{\mathrm{tc}}
=
\frac{1}{|\mathcal{B}|}
\sum_{i \in \mathcal{B}}
\beta_{m_i} c_i.
\]
This per-modality control prevents one teacher from dominating the shared backbone update and allows each modality to keep its own pace during distillation.

\subsection{Verification Reward}

Beyond token-level guidance, the routed teacher can also assess the quality of a complete rollout. We therefore reuse the same modality teacher as a verifier and add a reward that evaluates both the final answer and the reasoning trace produced by the student.

Let $a_i$ be the final answer extracted from $y_i$, $a_i^\star$ be the gold answer, and $S_i$ denote the token positions of the final answer. We first measure whether the routed teacher supports the student's final answer:
\[
A_i
=
\frac{1}{|S_i|}
\sum_{t \in S_i}
\log \pi_T^{m_i}(y_{i,t} \mid x_i, y_{i,<t}).
\]
To avoid rewarding confident but wrong answers, this term is gated by final correctness:
\[
\tilde{A}_i
=
\mathbf{1}[a_i=a_i^\star]\cdot \mathrm{clip}(A_i,-C,0).
\]
Here $C>0$ is the clipping threshold shared with the reasoning-gain term below.

We further measure whether the student's reasoning trace helps the teacher infer the correct answer. Let $z_i$ be the reasoning trace before the final answer. The reasoning gain is
\[
B_i
=
\log \pi_T^{m_i}(a_i^\star \mid x_i, z_i)
-
\log \pi_T^{m_i}(a_i^\star \mid x_i).
\]
Unlike the answer-confidence term, this gain is not gated by final correctness. Thus, difficult samples can still provide learning signal when the reasoning trace moves the teacher toward the correct answer, even if the final answer is wrong. We clip it as
\[
\tilde{B}_i = \mathrm{clip}(B_i,-C,C).
\]
The process reward is then
\begin{equation}
\label{eq:process-reward}
r_i^{\mathrm{proc}}
=
w_A \tilde{A}_i + w_B \tilde{B}_i,
\end{equation}
where $w_A$ and $w_B$ control the relative strength of answer confidence and reasoning gain.

\subsection{Training Objective}

For each rollout, we combine the task reward with the verification reward:
\[
r_i = r_i^{\mathrm{task}} + r_i^{\mathrm{proc}}.
\]
For each rollout $i$, let $\mathcal{G}_i\subseteq\mathcal{B}$ contain the rollouts sampled from the same prompt. Let $\mu_i$ and $\sigma_i$ denote the mean and standard deviation of $\{r_j:j\in\mathcal{G}_i\}$, respectively. We compute the rollout-level advantage as
\[
\hat{R}_i
=
\frac{r_i-\mu_i}{\sigma_i+\delta},
\]
where $\delta>0$ is a numerical stabilizer. We then optimize the student with an on-policy objective
\[
\mathcal{L}_{\mathrm{rl}}
=
-
\frac{1}{|\mathcal{B}|}
\sum_{i\in\mathcal{B}}
\hat{R}_i
\frac{1}{T_i}
\sum_{t=1}^{T_i}
\log \pi_\theta(y_{i,t}\mid x_i,y_{i,<t}).
\]
Let $h_{i,t}=(x_i,y_{i,<t})$ and let $\pi_{\mathrm{ref}}$ be the fixed policy before on-policy training. We use the rollout-averaged reference KL
\[
\mathcal{L}_{\mathrm{ref}}
=
\frac{1}{|\mathcal{B}|}
\sum_{i\in\mathcal{B}}
\frac{1}{T_i}
\sum_{t=1}^{T_i}
D_{\mathrm{KL}}\!\left(
\pi_\theta(\cdot\mid h_{i,t})
\,\|\,
\pi_{\mathrm{ref}}(\cdot\mid h_{i,t})
\right).
\]
The overall objective combines the policy loss, the routed teacher constraint, and this reference regularizer:
\begin{equation}
\label{eq:final-objective}
\mathcal{L}
=
\mathcal{L}_{\mathrm{rl}}
+
\mathcal{L}_{\mathrm{tc}}
+
\alpha \mathcal{L}_{\mathrm{ref}}.
\end{equation}
Here $\alpha\geq 0$ controls the reference KL strength. All three terms are computed from the same on-policy batch. The task and verification rewards form the rollout-level advantage optimized by $\mathcal{L}_{\mathrm{rl}}$, whereas $\mathcal{L}_{\mathrm{tc}}$ supplies dense token guidance only where the routed teacher is ahead. The reference term anchors the student to its initial policy. Gradients update only the shared student; the routed teachers and reference policy remain fixed. These terms therefore couple outcome, process, and token-level supervision on student-generated trajectories without changing the inference architecture.

\newcommand{\mainresultstable}{%
\begin{table*}[!ht]
\centering
\small
\renewcommand{\arraystretch}{1.08}
\setlength{\tabcolsep}{1.05pt}
\setlength{\aboverulesep}{0pt}
\setlength{\belowrulesep}{0pt}
\setlength{\extrarowheight}{0.5pt}
\resizebox{\textwidth}{!}{%
\begin{tabular*}{1.08\textwidth}{@{\extracolsep{\fill}}l*{5}{c}@{\hspace{4pt}}*{4}{c}@{\hspace{4pt}}*{2}{c}@{\hspace{4pt}}c@{\hspace{8pt}}>{\columncolor{overallblue}[0pt][0pt]}c@{}}
\toprule
\tworowhead{Model} & \multicolumn{5}{c}{\makebox[0pt][l]{\hspace{-1pt}Text}\phantom{Text}} & \multicolumn{4}{c}{\makebox[0pt][l]{\hspace{-2pt}Vision}\phantom{Vision}} & \multicolumn{2}{c}{\makebox[0pt][l]{\hspace{-2pt}Audio}\phantom{Audio}} & Omni & \textbf{Overall} \\
\cmidrule(l{4pt}r{6pt}){2-6}\cmidrule(l{3pt}r{7pt}){7-10}\cmidrule(l{3pt}r{7pt}){11-12}\cmidrule(l{1pt}r{9pt}){13-13}\cmidrule(l{0pt}r{0pt}){14-14}
& \benchhead{AIME25} & \benchhead{AIME26} & \benchhead{HQA} & \benchhead{MMLU-Pro} & \benchhead{GPQA} & \benchhead{MMMU} & \benchhead{MathV.} & \benchhead{ChartQA} & \benchhead{AOKVQA} & \benchhead{MMAU} & \benchhead{AVQA} & \benchhead{OmniBench} & \textbf{Avg.} \\
\midrule
\multicolumn{13}{l}{\textit{Qwen3-Omni-30B-A3B}} & \phantom{0} \\
Text teacher  & 47.9 & 57.7 & \textbf{41.5} & 76.8 & \underline{65.8} & 67.7 & 76.3 & 86.5 & 87.7 & 76.0 & 79.8 & \underline{61.1} & \underline{68.7} \\
Image teacher & 47.9 & 52.1 & 27.3 & 77.3 & 64.8 & 68.2 & \underline{77.3} & \textbf{87.8} & 88.6 & 75.7 & 80.8 & \textbf{61.5} & 67.4 \\
Audio teacher & \underline{52.1} & 59.2 & 27.3 & 76.9 & 64.0 & 67.4 & 76.4 & 87.2 & \underline{88.9} & \textbf{79.1} & \textbf{84.4} & 55.6 & 68.2 \\
Base          & 49.2 & 52.9 & 26.8 & 77.0 & 65.3 & 67.9 & 76.3 & 85.8 & 87.6 & 75.6 & 79.3 & 60.3 & 67.0 \\
GRPO          & 49.2 & 53.8 & 28.0 & 76.7 & 65.3 & 67.9 & 76.7 & 86.2 & 87.9 & 76.2 & 80.5 & 59.7 & 67.3 \\
Native OPD    & 51.7 & 55.8 & \underline{41.1} & \underline{77.5} & 64.9 & \underline{68.4} & 76.4 & 73.4 & 88.0 & 75.8 & 81.2 & 60.3 & 67.9 \\
ExOPD         & 50.0 & \textbf{64.2} & 41.0 & 77.1 & 63.5 & 68.0 & 76.5 & 87.2 & 88.7 & 65.1 & \underline{83.2} & 58.8 & 68.6 \\
\ours{}       & \textbf{55.0} & \underline{63.3} & \underline{41.1} & \textbf{78.2} & \textbf{68.7} & \textbf{68.7} & \textbf{77.6} & \underline{87.6} & \textbf{89.1} & \underline{76.6} & 82.6 & \underline{61.1} & \textbf{70.8} \\
\midrule
\multicolumn{13}{l}{\textit{Qwen2.5-Omni-7B}} & \phantom{0} \\
Base          & 5.0 & 3.3 & 24.1 & 32.5 & 31.8 & 51.7 & \underline{67.5} & 66.2 & 85.1 & 71.4 & 76.8 & 47.2 & 46.9 \\
GRPO          & \underline{10.0} & 3.3 & 22.9 & 38.5 & \underline{32.1} & 51.4 & \textbf{67.7} & 67.2 & \textbf{86.5} & 73.0 & \textbf{80.3} & 44.8 & 48.1 \\
Native OPD    & 6.7 & 3.3 & 28.8 & \underline{52.4} & \textbf{32.3} & \underline{53.1} & 66.8 & 69.1 & 81.7 & \underline{73.1} & \underline{80.0} & \underline{47.3} & 49.5 \\
ExOPD         & 6.7 & \textbf{8.3} & \underline{29.5} & \underline{52.4} & 30.8 & 51.4 & 66.4 & \underline{73.8} & 82.6 & 72.0 & 78.4 & 46.3 & \underline{49.9} \\
\ours{}       & \textbf{11.7} & \underline{6.7} & \textbf{29.8} & \textbf{52.5} & 31.1 & \textbf{54.0} & \underline{67.5} & \textbf{79.7} & \underline{85.7} & \textbf{75.0} & 79.1 & \textbf{48.0} & \textbf{51.7} \\
\midrule
\multicolumn{13}{l}{\textit{Qwen2.5-Omni-3B}} & \phantom{0} \\
Base          & 1.7 & 1.7 & 18.1 & 22.8 & 24.5 & 42.8 & 53.8 & 57.3 & 81.9 & 69.9 & 72.4 & 41.2 & 40.7 \\
GRPO          & \underline{3.3} & 1.7 & 17.4 & 24.5 & \underline{29.0} & 44.6 & 56.4 & 59.7 & \underline{83.8} & \textbf{71.6} & 74.9 & \underline{41.7} & 42.4 \\
Native OPD    & 1.7 & \underline{3.3} & 23.5 & 36.4 & 26.5 & \underline{44.9} & 59.4 & 61.2 & \textbf{84.3} & 69.3 & 75.4 & \textbf{42.6} & 44.0 \\
ExOPD         & 1.7 & \underline{3.3} & \underline{24.5} & \underline{37.9} & \textbf{30.3} & 43.9 & \underline{59.9} & \underline{62.3} & 83.6 & 69.9 & \underline{76.3} & 40.2 & \underline{44.5} \\
\ours{}       & \textbf{5.0} & \textbf{6.7} & \textbf{25.6} & \textbf{41.5} & 28.5 & \textbf{45.8} & \textbf{60.5} & \textbf{67.5} & 83.7 & \underline{70.8} & \textbf{76.9} & 41.5 & \textbf{46.2} \\
\bottomrule
\end{tabular*}
}
\caption{Main results across text, visual, audio, and omni-modal benchmarks. All numbers are accuracy in percentage. For each column within a backbone block, the best result is shown in \textbf{bold} and the second-best result is \underline{underlined}; specialist teachers are included in the upper-block comparison.}
\label{tab:main_results}
\end{table*}
}

\section{Experiments}

\subsection{Experimental Setup}

\paragraph{Models and teachers.}
Our main experiments use Qwen3-Omni-30B-A3B-Instruct~\citep{xu2025qwen3omni} as the student backbone. Starting from the same backbone, we construct three modality-specialized teachers by applying GRPO separately to the corresponding text, image, and audio training data. During distillation, each student rollout is routed to the teacher that matches the modality of the prompt. We also evaluate smaller Qwen2.5-Omni students~\citep{qwen25omni2025} to test whether the same recipe transfers to dense omni-modal models with a different multimodal tokenizer and processor.

\paragraph{Training data.}
All students and teachers are trained on the same modality-balanced prompt set covering text, image, and audio reasoning, drawn from public reasoning and multimodal question-answering sources with verifiable answers. The training prompts are disjoint from every evaluation benchmark: we remove exact and normalized-string overlaps against all test sets, so no benchmark question appears during training. The full source list, per-modality composition, filtering, and licenses are given in the supplementary material.

\paragraph{Baselines.}
We compare \ours{} with several baselines. The base model is the original omni-modal instruction model before further training. GRPO~\citep{shao2024deepseekmath} trains one student on the union of the text, image, and audio data using task rewards only, without teacher distillation. Native OPD applies standard on-policy distillation from the routed modality teacher~\citep{agarwal2024onpolicy,gu2024minillm}. ExOPD follows reward extrapolation and uses a stronger teacher direction to encourage improvement beyond the teacher~\citep{yang2026learning}. We also report the three separately GRPO-trained modality specialists as references rather than as deployable ensemble models.

\paragraph{Benchmarks.}
We evaluate text, visual, audio, and omni-modal reasoning. The text benchmarks include AIME25, AIME26, HotpotQA~\citep{yang2018hotpotqa}, MMLU-Pro~\citep{wang2024mmlupro}, and GPQA~\citep{rein2023gpqa}. The visual benchmarks include MMMU~\citep{yue2024mmmu}, MathVista~\citep{lu2024mathvista}, ChartQA~\citep{masry2022chartqa}, and A-OKVQA~\citep{schwenk2022okvqa}. The audio benchmarks include MMAU~\citep{sakshi2025mmau} and AVQA~\citep{yang2022avqa}, and we use OmniBench~\citep{li2026omnibench} for omni-modal evaluation. All results are reported as accuracy, and the overall score is the average over these twelve benchmarks.

\paragraph{Implementation details.}
We train on a cluster of 32 NVIDIA H20 GPUs. Unless otherwise specified, student training uses 16 GPUs, while the remaining GPUs host the text, audio, and image teacher servers for online guidance. We use eight samples per prompt, a rollout batch size of 16, and a global batch size of 64. The maximum prompt length and the maximum response length are both 8,192 tokens. For \ours{}, we use the verification reward with $w_A=0.2$, $w_B=0.2$, and clipping threshold $C=2.0$. The dual controller estimates each $\epsilon_m$ from a 10-step warm-up and uses $\epsilon_{\min}=0.02$. We keep a reference KL regularizer in all on-policy runs to reduce policy drift. All baselines share the same backbone, data, batching, and evaluation protocol, differing only in their guidance objective, and use a single run per configuration with the same random seed. Full hyperparameters, search ranges, and the software and hardware environment are listed in the supplementary material, and the code is publicly available through the link on the first page.

\newcommand{\ablationresultstable}{%
\begin{table*}[!ht]
\centering
\footnotesize
\renewcommand{\arraystretch}{1.04}
\setlength{\tabcolsep}{1.10pt}
\resizebox{0.97\textwidth}{!}{%
\begin{tabular}{@{}l*{5}{c}@{\hspace{4pt}}*{4}{c}@{\hspace{4pt}}*{2}{c}@{\hspace{4pt}}c@{\hspace{8pt}}c@{\hspace{5pt}}c@{\hspace{6pt}}}
\toprule
\tworowhead{Variant} & \multicolumn{5}{c}{Text} & \multicolumn{4}{c}{Vision} & \multicolumn{2}{c}{Audio} & Omni & \multicolumn{2}{c}{\textbf{Overall}} \\
\cmidrule(lr){2-6}\cmidrule(lr){7-10}\cmidrule(lr){11-12}\cmidrule(lr){13-13}\cmidrule(lr){14-15}
& \benchhead{AIME25} & \benchhead{AIME26} & \benchhead{HQA} & \benchhead{MMLU-Pro} & \benchhead{GPQA} & \benchhead{MMMU} & \benchhead{MathV.} & \benchhead{ChartQA} & \benchhead{AOKVQA} & \benchhead{MMAU} & \benchhead{AVQA} & \benchhead{OmniBench} & \makebox[2.5em][c]{\textbf{Avg.}} & \makebox[2.0em][c]{$\mathbf{\Delta}$} \\
\midrule
Full \ours{}             & 55.0 & 63.3 & 41.1 & 78.2 & 68.7 & 68.7 & 77.6 & 87.6 & 89.1 & 76.6 & 82.6 & 61.1 & 70.8 & \textemdash \\
\addlinespace[1.5pt]
\quad$-$ One-Sided Guidance   & 50.0 & 60.0 & 40.5 & 77.0 & 63.4 & 68.2 & 76.7 & 87.2 & 87.7 & 76.3 & 80.9 & 60.3 & 69.0 & $-1.8$ \\
\quad$-$ Modality Control     & 47.1 & 60.4 & 36.9 & 77.6 & 65.6 & 67.9 & 76.0 & 86.5 & 88.5 & 76.0 & 80.3 & 60.5 & 68.6 & $-2.2$ \\
\quad$-$ Verification Reward & 48.3 & 60.0 & 40.6 & 77.3 & 65.9 & 66.9 & 76.1 & 86.5 & 87.9 & 75.3 & 80.8 & 59.7 & 68.8 & $-2.0$ \\
\quad$-$ Reasoning Gain       & 52.9 & 63.3 & 38.8 & 77.1 & 65.0 & 67.7 & 76.4 & 85.8 & 88.0 & 76.0 & 80.1 & 59.1 & 69.2 & $-1.6$ \\
\quad$-$ Answer Confidence    & 51.7 & 60.0 & 40.9 & 77.1 & 64.1 & 68.1 & 76.8 & 87.8 & 88.6 & 76.0 & 81.7 & 59.8 & 69.4 & $-1.4$ \\
\bottomrule
\end{tabular}
}%
\caption{Ablation results on Qwen3-Omni-30B-A3B. $\Delta$ denotes the change in Avg. relative to Full \ours{}.}
\label{tab:ablation}
\end{table*}
}

\subsection{Main Results}

\paragraph{Overall performance.}
Table~\ref{tab:main_results} reports the main evaluation results. On Qwen3-Omni-30B-A3B, \ours{} achieves the best overall average of 70.8, outperforming the base model, mixed-modality GRPO, native OPD, ExOPD, and the strongest modality-specialized teacher by 3.8, 3.5, 2.9, 2.2, and 2.1 points, respectively, while retaining a single deployable model. It surpasses both the base model and GRPO on all 12 benchmarks, improves over native OPD on 11 while tying the remaining one, and exceeds ExOPD on 10. When the specialist teachers are included, \ours{} ranks first or second on 11 of 12 benchmarks.

\mainresultstable
\FloatBarrier

\paragraph{Specialist consolidation.}
The 30B result goes beyond selecting the best teacher benchmark by benchmark. Compared separately with the text, image, and audio teachers, \ours{} scores higher on 10 of 12 benchmarks against each specialist. Moreover, an oracle that selects the strongest teacher separately for every benchmark reaches an average of 70.3, still 0.5 points below the single \ours{} student. Surpassing this oracle shows that joint consolidation recovers improvements beyond simply preserving the strongest specialist for each benchmark.

\paragraph{Cross-modal balance.}
More importantly, \ours{} improves all modality groups rather than trading one capability for another. Relative to the base model, its aggregated text, vision, audio, and omni-modal scores increase by 7.0, 1.4, 2.2, and 0.8 points, respectively. In contrast, GRPO drops by 0.6 points on omni-modal evaluation, native OPD loses 2.9 points on vision, and ExOPD loses 3.3 points on audio and 1.5 points on omni-modal evaluation. These results support the need to control teacher guidance separately across modalities when training a shared omni-modal backbone.

\begin{wrapfigure}{l}{0.49\textwidth}
\vspace{-0.6\baselineskip}
\centering
\includegraphics[width=\linewidth]{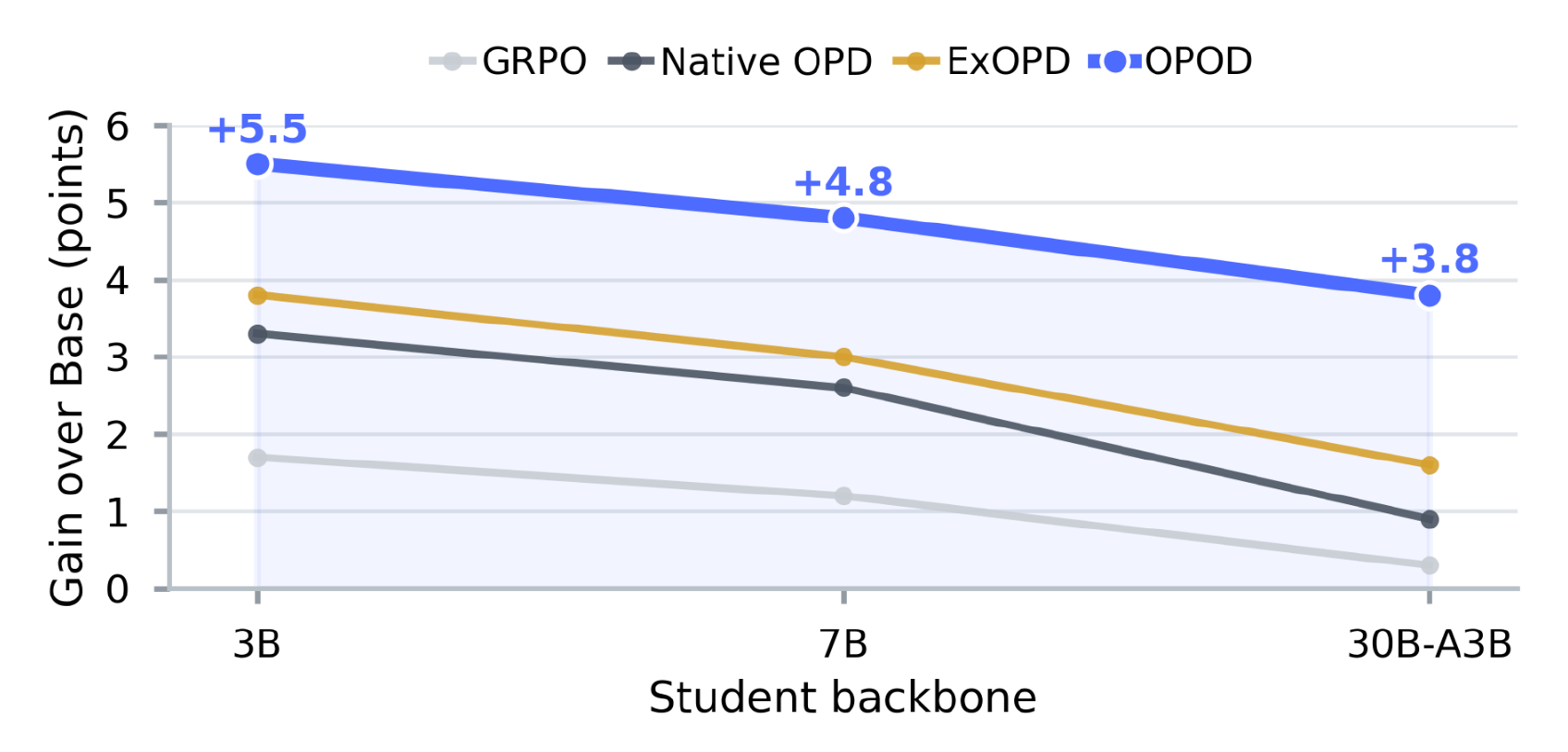}
\Description{A bar chart comparing average benchmark improvements over each base model for GRPO, native OPD, ExOPD, and OPOD at 3B, 7B, and 30B scales.}
\caption{Average improvement over the corresponding base model. \ours{} provides the largest gain at every scale.}
\label{fig:scale_comparison}
\vspace{-0.4\baselineskip}
\end{wrapfigure}

\paragraph{Scaling across student backbones.}
\mbox{Figure~\ref{fig:scale_comparison}} summarizes the average improvement over the corresponding base model at each scale. \ours{} delivers the largest gain throughout, improving the 3B, 7B, and 30B-A3B students by 5.5, 4.8, and 3.8 points, respectively. At 7B and 3B, it ranks first or second on 10 and 9 benchmarks; its largest base-relative gains span MMLU-Pro (20.0 and 18.7 points) and ChartQA (13.5 and 10.2). This consistency shows that the method transfers across dense and mixture-of-experts omni-modal backbones.

The comparison also separates general post-training headroom from method-specific gains. Improvements over the base become smaller as backbone capacity increases for every baseline, yet \ours{} retains a 3.8-point gain on 30B-A3B, compared with 1.6 points for ExOPD, 0.9 for native OPD, and 0.3 for GRPO. Consequently, its margin over the strongest alternative grows from 1.7 points at 3B to 1.8 at 7B and 2.2 at 30B-A3B. The advantage therefore persists even when a stronger base leaves less room for generic post-training improvements.

The method ordering is unchanged across scales, and \ours{} maintains margins of 3.8, 3.6, and 3.5 points over mixed-modality GRPO, reinforcing that its gain is not tied to one parameter regime.

\FloatBarrier
\subsection{Ablation Study}

\paragraph{Core components.}
Table~\ref{tab:ablation} reports the Qwen3-Omni-30B-A3B results. Removing one-sided guidance, adaptive modality control, or the verification reward lowers the average by 1.8, 2.2, and 2.0 points, respectively. Across the resulting 60 benchmark--variant comparisons, Full \ours{} matches or exceeds the corresponding ablation in 59. The improvements therefore recur across benchmarks rather than arising from a small set of favorable tasks.

\ablationresultstable
\FloatBarrier

\paragraph{Modality control.}
Replacing the three adaptive weights $\beta_m$ with one fixed weight shared across modalities produces the largest degradation of 2.2 points. The corresponding text, vision, audio, and omni-modal group scores decrease by 3.7, 1.0, 1.5, and 0.6 points, respectively. The decline in every evaluation group shows that a shared weight creates broad cross-modal tradeoffs rather than an isolated benchmark failure. Text drops most even though image retains guidance longest in Figure~\ref{fig:experiment}, confirming that guidance duration reflects calibrated constraint gaps rather than modality difficulty.

\paragraph{Reward decomposition.}
We further separate the two terms in the verification reward. Removing reasoning gain and answer confidence reduces the average by 1.6 and 1.4 points, respectively. Full \ours{} matches or exceeds the variant without reasoning gain on every benchmark and outperforms the variant without answer confidence on 11 of 12. Their profiles also differ: reasoning gain has larger effects on AVQA and OmniBench, whereas answer confidence matters more on AIME25, AIME26, and GPQA. The former evaluates whether the generated reasoning increases support for the gold answer, while the latter measures teacher support for a correct final answer. The two signals therefore capture complementary aspects of trajectory quality.

\subsection{Analysis of Adaptive Modality Control}

\begin{figure}[!t]
\centering
\includegraphics[width=\columnwidth]{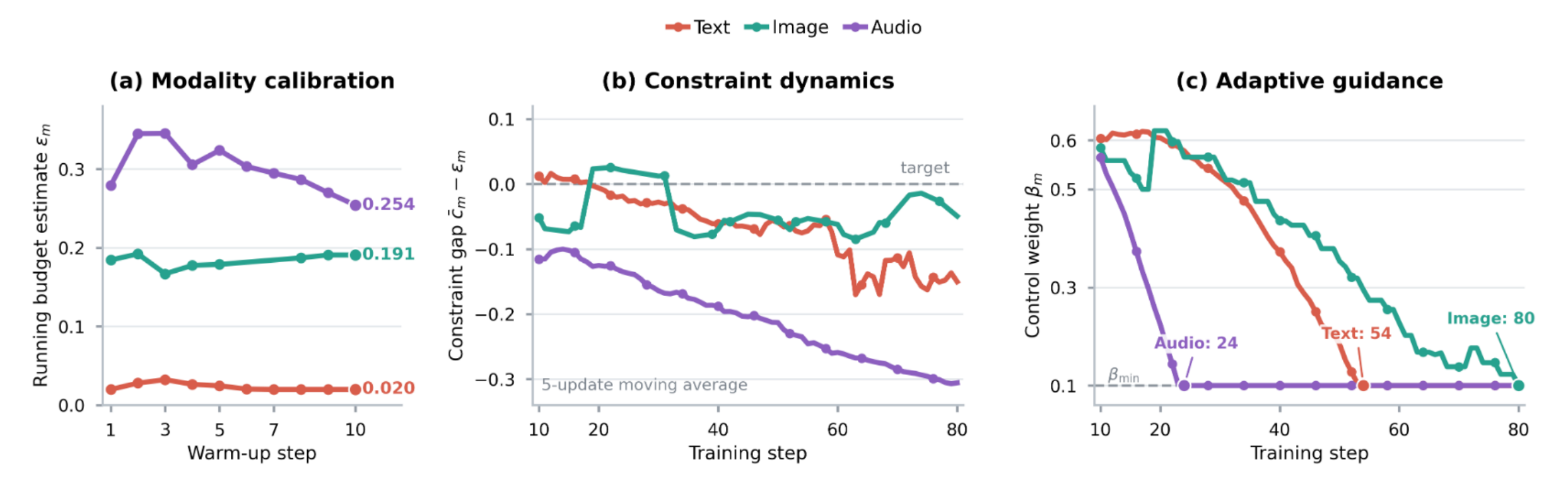}
\Description{Three plots show modality-specific constraint budgets, constraint gaps, and adaptive control weights over training for text, image, and audio.}
\caption{Dynamics of adaptive modality control in the Qwen3-Omni-30B-A3B \ours{} run. (a) Running estimates of the modality-specific constraint budgets during the 10-step warm-up, yielding frozen values of $0.020$, $0.191$, and $0.254$ for text, image, and audio. (b) Five-update moving averages of the constraint gaps $\bar{c}_m-\epsilon_m$, where zero denotes each modality's calibrated target. (c) The corresponding control weights $\beta_m$. Annotations mark the first optimization step at which each weight reaches $\beta_{\min}=0.1$.}
\label{fig:experiment}
\end{figure}

Adaptive modality control produces the largest ablation drop. We therefore examine its statistics in the full 30B \ours{} run over the 10-step warm-up and the first 80 optimization steps (Figure~\ref{fig:experiment}). Because the weights are updated from observed constraint gaps, their trajectories show whether teacher guidance is actually decoupled across modalities.

\paragraph{Modality-specific calibration.}
Figure~\ref{fig:experiment}(a) yields frozen budgets of $\epsilon_{\mathrm{text}}=0.020$, $\epsilon_{\mathrm{image}}=0.191$, and $\epsilon_{\mathrm{audio}}=0.254$. Their $12.7\times$ range shows that the constraints operate at substantially different scales across modalities. These budgets characterize the natural scale of teacher--student discrepancy during warm-up rather than the intrinsic difficulty of a modality. A shared target would therefore overconstrain some modalities while underconstraining others.

\paragraph{Data-driven control trajectories.}
After calibration, Figure~\ref{fig:experiment}(b) shows the five-update moving average of each constraint gap $\bar{c}_m-\epsilon_m$. A positive gap indicates that the discrepancy exceeds its calibrated budget and preserves or increases the corresponding teacher weight, whereas a negative gap allows the controller to reduce it. Audio remains below its target throughout the plotted window, text gradually moves farther below its target, and image stays closer to its boundary for longer. Consequently, although all three weights begin near $0.6$, they first reach $\beta_{\min}=0.1$ at different optimization steps: 24 for audio, 54 for text, and 80 for image (Figure~\ref{fig:experiment}(c)). These arrival times indicate when each modality enters its own calibrated budget; they should not be interpreted as a ranking of modality difficulty. The weights are therefore neither shared nor governed by a prescribed common schedule. Instead, the controller releases teacher pressure for modalities within budget while retaining guidance longer for those near their calibrated boundaries.

\FloatBarrier
\section{Conclusion}

We presented \ours{}, an on-policy omni distillation framework that consolidates text, image, and audio specialists into one omni-modal policy. It combines one-sided token guidance that avoids a teacher ceiling, modality-specific control over separately calibrated budgets, and teacher verification of complete trajectories. Together, these components coordinate conflicting guidance while allowing the student to move beyond its teachers. Across twelve benchmarks, \ours{} achieves the best overall average on all three backbones, outperforming the strongest comparator by 2.1, 1.8, and 1.7 points on 30B, 7B, and 3B. Ablations validate each component, while discarding the teachers after training preserves single-student inference during deployment.

\bibliographystyle{assets/plainnat}
\bibliography{citation}

\end{document}